\documentclass[10pt,twocolumn,letterpaper]{article}

\usepackage[top=0.8in,bottom=0.85in,left=0.72in,right=0.72in,columnsep=0.28in]{geometry}
\usepackage[sc]{mathpazo}
\usepackage[T1]{fontenc}
\usepackage{amsmath,amssymb}
\usepackage{graphicx}
\usepackage{booktabs}
\usepackage{multirow}
\usepackage{xcolor}
\usepackage[expansion=false]{microtype}
\usepackage{url}
\usepackage[font=small,labelfont=bf,skip=5pt]{caption}
\usepackage{subcaption}
\usepackage{titlesec}
\usepackage{enumitem}
\usepackage{tikz}
\usetikzlibrary{positioning,fit,arrows.meta,calc}
\usepackage[colorlinks=true,linkcolor=blue!55!black,citecolor=blue!55!black,urlcolor=blue!55!black]{hyperref}

\setlist{itemsep=1pt,topsep=3pt,parsep=0pt,leftmargin=1.2em}
\titleformat{\section}{\large\bfseries}{\thesection.}{0.5em}{}
\titleformat{\subsection}{\normalsize\bfseries}{\thesubsection}{0.5em}{}
\titlespacing*{\section}{0pt}{9pt plus 2pt minus 2pt}{4pt plus 1pt}
\titlespacing*{\subsection}{0pt}{7pt plus 2pt minus 1pt}{3pt plus 1pt}
\title{\vspace{-1.4em}\Large\textbf{PAWP-TTS: Phoneme-Aware Weight Programming for\\ Compact Bilingual English--Nepali Speech Synthesis}}

\author{
  \textbf{Firoj Paudel} \\
  \texttt{firojpaudel@gmail.com} \\[3pt]
  \small Code: \url{https://github.com/Firojpaudel/PAWP_TTS}\quad
  Audio: \url{https://huggingface.co/spaces/Firoj112/pawp-tts-demo}
}

\date{}

\begin{document}

\twocolumn[
  \maketitle
  \vspace{-1.6em}
  \begin{center}
    \begin{minipage}{0.88\textwidth}
      \small
\noindent\textbf{Abstract.}\;
This paper presents \textbf{PAWP-TTS}, a compact non-autoregressive text-to-speech system for English, Nepali, and English--Nepali code-mixed speech at 24\,kHz, with no language model at inference. Deterministic pronunciation knowledge lives in a rule- and lexicon-based bilingual frontend that preserves native Nepali contrasts and avoids Hindi phonology, so the network only models acoustics. Counting every learned inference tensor, including the waveform decoder, \textbf{PAWP-1.5M} (1,490,390 parameters, 5.98\,MB) reaches 1.50\% English and 1.75\% Nepali character error rate under a single ASR recognizer on small held-out sets, and 5.04\% / 4.02\% Nepali / English phoneme error rate (PER) on 600 untouched test utterances. Because these metrics are partly circular, all systems are re-scored with two independent recognizers. Under these, synthetic speech is close to, but no better than, human recordings (English phone error 15.1\% vs.\ 13.4\%), and PAWP-1.5M reaches 18.1\% WER on a new 160-sentence code-mixed set. No listening test has been run yet. Decoder-centric \textbf{281K-parameter} models (1.1\,MB) remain intelligible, and at equal unique parameters a dense model has lower phone error than cross-layer basis sharing on all 6 seed$\times$language pairs under the independent recognizer ($p=0.01$). Fixing a frontend/label version mismatch removed a spurious Nepali substitution error entirely. The paper then studies \emph{weight-space programming}: inspection-guided, zero-training weight edits. Under strict controls (validation-only selection, one-shot test, paired bootstrap, matched random and same-parameter gradient baselines), edits show real selectivity (27:1 on word-boundary pauses with 96 weights) but no accuracy gain beyond controls in this exploratory study. Two apparent targets proved to be label or phonology artifacts. Code, ONNX graphs and weights are released.
\normalsize

    \end{minipage}
  \end{center}
  \vspace{1.0em}
]

\section{Introduction}
\label{sec:intro}

Recent neural text-to-speech (TTS) systems increasingly rely on hundred-million to billion-parameter language models and neural audio codecs to achieve natural prosody and multilingual coverage. While effective in cloud deployments, these architectures incur high memory footprints, large time-to-first-audio (TTFA), and poor accessibility for low-resource languages such as Nepali---especially in edge, CPU-only, or browser-targeted settings.

Nepali presents specific phonological challenges that are routinely mishandled by generic Indic or Hindi-biased multilingual models~\cite{khatiwada2009nepali}: standard Kathmandu Nepali maintains a four-way stop laryngeal contrast, dental versus retroflex stops with post-vocalic flapping, alveolar affricates (/ts, ts$^h$, dz, dz$^h$/) rather than Hindi postalveolars, contrastive consonant gemination, an open-mid back unrounded inherent vowel rather than the Hindi central schwa, and an alveolar sibilant merger (/s/) across orthographic \textit{sa/\'{s}a/\d{s}a}. Moreover, everyday Nepali speech frequently code-switches mid-sentence with English.

This work investigates two research questions:
\begin{enumerate}
  \item \textbf{Compact Bilingual Capacity:} How small can a complete, end-to-end 24\,kHz bilingual English--Nepali TTS system (including its neural waveform decoder) be made when deterministic linguistic knowledge is cleanly separated into a rule- and lexicon-backed frontend?
  \item \textbf{Diagnostic Weight-Space Programming (WSP):} Following the weight-space programming view that trained weights can be read and edited like a program~\cite{wsp2026}, can inspection-guided, zero-training edits (using rank-one tools such as ROME~\cite{meng2022rome}) surgically repair phonetic or timing defects in a compact TTS model without degrading the other language, and do such edits hold up against matched random and same-parameter gradient controls?
\end{enumerate}

The contributions are:
\begin{itemize}
  \item \textbf{Open Compact Bilingual TTS at 1.5M and 281K Parameters:} Complete parameter accounting (unique stored vs.\ materialized parameters and serialized size, every inference tensor including the decoder; frontend lexicon and \texttt{libespeak-ng} reported separately) across \texttt{PAWP-1.5M} (1,490,390 params, 5.98\,MB FP32) and \texttt{PAWP-300K} (281,078 unique / 328,158 materialized params, 1.12\,MB FP32) with language-gated low-rank residuals that fold statically into zero-overhead monolingual ONNX graphs.
  \item \textbf{Clean Frontend--Alignment Consistency:} Empirical demonstration that eliminating a frontend/cache version mismatch (\texttt{v2} $\rightarrow$ \texttt{v3}) nearly halves validation Nepali PER (7.06\% $\rightarrow$ 3.96\%) and eliminates a 55.1\% spurious \textit{va} $\rightarrow$ \texttt{/w/} artifact.
  \item \textbf{A Controlled Audit of Zero-Training WSP in Compact TTS:} Experiments with validation-only selection, one-shot tests and paired-bootstrap 95\% CIs show real selectivity (27:1 pause selectivity in 96 weights; English left untouched by construction when editing Nepali-gated residuals) but \emph{no accuracy gain beyond matched controls}, and identify three confounders that mimic model defects: cache-label drift, native phonology mismatch, and global duration bias.
\end{itemize}

\section{Related Work}
\label{sec:related}

\paragraph{Non-Autoregressive TTS and Fourier Vocoding.}
The acoustic backbone builds on non-autoregressive variance-adapted synthesis~\cite{ren2021fastspeech2} paired with Fourier-domain ConvNeXt1D + iSTFT waveform synthesis~\cite{siuzdak2024vocos}. Unlike conventional pipelines that exclude a multi-million-parameter vocoder from reported parameter counts, PAWP-TTS counts every learned tensor required for inference from phoneme IDs to 24\,kHz PCM waveforms. Compact TTS work has reduced acoustic models through architecture search~\cite{luo2021lightspeech} and module-wise distillation~\cite{chevi2022nixtts}; PAWP-TTS differs by also counting the waveform decoder and targeting a bilingual low-resource language. Articulatory feature conditioning follows prior low-resource multilingual TTS work~\cite{lux2022toucan}. Phoneme durations are commonly obtained by monotonic alignment search~\cite{kim2020glowtts} or jointly learned aligners~\cite{badlani2022onealign}; on the mixed multi-speaker bilingual data used here a jointly learned aligner collapsed, so CTC~\cite{graves2006ctc} forced alignment is used instead (Section~\ref{sec:exp}).

\paragraph{Model Compression and Parameter Sharing.}
Prior compression methods include Fisher-weighted SVD~\cite{hsu2022fwsvd}, adaptive low-rank allocation~\cite{zhang2023adalora}, dynamic sparse training~\cite{evci2020rigl}, and cross-layer basis sharing~\cite{wang2025basissharing}. This work explicitly distinguishes \emph{unique stored parameters} from \emph{materialized compute parameters} when evaluating cross-layer shared bases against equal-unique-parameter dense baselines.

\paragraph{Mechanistic Localization and Weight Editing.}
Rank-one model editing with covariance whitening (ROME)~\cite{meng2022rome} demonstrated targeted, gradient-free edits in language models, and weight-space programming~\cite{wsp2026} frames such edits as reading and rewriting a network's weights like a program, noting that edits can repair but not create capabilities. This work adapts the inspect--localize--edit--verify workflow to compact bilingual speech synthesis, imposing strict evaluation controls: validation-only hyperparameter selection, one-shot evaluation on untouched test data, paired-bootstrap confidence intervals, matched-norm random edits, and same-parameter gradient fine-tuning controls.

\section{Frontend and Compact Architecture}
\label{sec:arch}

\begin{figure*}[t]
  \centering
  \resizebox{\textwidth}{!}{
\begin{tikzpicture}[
  font=\footnotesize,
  >={Stealth[length=4pt,width=3.2pt]},
  blk/.style={draw=black!75, line width=0.5pt, rounded corners=2pt, minimum height=5.6mm,
              minimum width=30mm, align=center, inner sep=2pt},
  enc/.style={blk, fill=blue!7},
  var/.style={blk, fill=green!8},
  aco/.style={blk, fill=violet!7},
  dec/.style={blk, fill=red!6},
  emb/.style={blk, fill=cyan!7},
  io/.style={blk, draw=none, fill=none, minimum height=4.6mm},
  det/.style={blk, dashed, fill=black!3},
  sm/.style={blk, minimum width=21mm, minimum height=5mm},
  op/.style={circle, draw=black!75, line width=0.5pt, inner sep=0pt, minimum size=3.6mm},
  rep/.style={font=\scriptsize\itshape, text=black!65},
  prm/.style={font=\scriptsize, text=black!55, anchor=west},
  flow/.style={->, line width=0.55pt, black!80},
  ttl/.style={font=\small\bfseries, anchor=south}
]

\begin{scope}[local bounding box=A]
  \node[io]  (txt)  at (0,0) {Text (EN / NE / code-mixed)};
  \node[det, above=2.6mm of txt] (fe) {Bilingual frontend (0 params)\\[-1pt]{\scriptsize phonemes + language, stress, boundary tags}};
  \node[emb, above=2.6mm of fe] (emb) {Embedding};
  \node[enc, above=3.2mm of emb, minimum height=7mm] (enc) {Gated ConvNeXt1D block $\times 4$};
  \node[var, above=3.4mm of enc] (dur) {Duration predictor};
  \node[var, above=2.4mm of dur] (lr)  {Length regulator};
  \node[var, above=2.4mm of lr]  (pe)  {Pitch / energy predictors};
  \node[aco, above=3.4mm of pe, minimum height=7mm] (aco) {Gated ConvNeXt1D block $\times 4$};
  \node[aco, above=2.4mm of aco] (mel) {Linear $\rightarrow$ log-mel};
  \node[dec, above=3.4mm of mel, minimum height=7mm] (dec) {iSTFT waveform decoder};
  \node[io,  above=2.6mm of dec] (wav) {24\,kHz waveform};

  \foreach \a/\b in {txt/fe, fe/emb, emb/enc, enc/dur, dur/lr, lr/pe, pe/aco, aco/mel, mel/dec, dec/wav}
    \draw[flow] (\a) -- (\b);

  \node[draw=black!45, dashed, rounded corners=3pt, fit=(dur)(lr)(pe), inner sep=2.2mm] (va) {};
  \node[rep, left=0.6mm of va.west, rotate=90, anchor=south] {variance adaptor};
  \node[prm] at ([xshift=2.2mm]emb.east) {18K};
  \node[prm] at ([xshift=2.2mm]enc.east) {367K};
  \node[prm] at ([xshift=2.2mm]va.east)  {62K};
  \node[prm] at ([xshift=2.2mm]aco.east) {378K};
  \node[prm] at ([xshift=2.2mm]dec.east) {665K};
  \node[font=\scriptsize, text=black!60, align=center] (mask) at ([xshift=-17mm]lr.west) {language\\mask $m(t)$};
  \draw[->, black!50, densely dotted, line width=0.5pt] (mask.south) |- (enc.west);
  \draw[->, black!50, densely dotted, line width=0.5pt] (mask.north) |- (aco.west);
  \draw[->, black!55, line width=0.45pt] (dur.east) -- ++(3mm,0) |- (lr.east);
\end{scope}
\node[ttl] at ([yshift=1mm]A.north) {(a) PAWP-1.5M (1,490,390 parameters)};

\begin{scope}[shift={(8.2,2.55)}, local bounding box=B]
  \node[io] (bx) at (0,0) {$x$};
  \node[enc, sm, above=3mm of bx] (dw) {DWConv1D (look-ahead 1)};
  \node[enc, sm, above=2.4mm of dw] (ln) {LayerNorm};
  \node[enc, sm, above=2.4mm of ln, minimum height=6.4mm] (pw1) {$W_1 x$};
  \node[enc, sm, above=2.4mm of pw1] (gl) {GELU};
  \node[enc, sm, above=2.4mm of gl, minimum height=6.4mm] (pw2) {$W_2 h$};
  \node[op, above=3mm of pw2] (add) {$+$};
  \node[io, above=2.4mm of add] (by) {$y$};
  \foreach \a/\b in {bx/dw, dw/ln, ln/pw1, pw1/gl, gl/pw2, pw2/add, add/by} \draw[flow] (\a) -- (\b);
  \draw[flow] ([yshift=1.2mm]bx.north) -- ++(19mm,0) |- (add.east);
  \node[font=\scriptsize, text=black!55, anchor=west] at ([xshift=13.8mm,yshift=-6mm]add.east -| bx.north) {};

  \node[blk, fill=orange!10, minimum width=13mm, minimum height=5mm, font=\scriptsize, anchor=east]
        (ren) at ([xshift=-5mm,yshift=3.4mm]pw1.west) {$U_{\mathrm{en}}V_{\mathrm{en}}$};
  \node[blk, fill=orange!10, minimum width=13mm, minimum height=5mm, font=\scriptsize, anchor=east]
        (rne) at ([xshift=-5mm,yshift=-3.4mm]pw1.west) {$U_{\mathrm{ne}}V_{\mathrm{ne}}$};
  \draw[black!70, line width=0.45pt] (ren.east) -- ++(1.6mm,0) |- ([yshift=1pt]pw1.west);
  \draw[black!70, line width=0.45pt] (rne.east) -- ++(1.6mm,0) |- ([yshift=-1pt]pw1.west);
  \node[font=\scriptsize, text=black!60, left=0.5mm of ren] {$m_{\mathrm{en}}(t)\,\cdot$};
  \node[font=\scriptsize, text=black!60, left=0.5mm of rne] {$m_{\mathrm{ne}}(t)\,\cdot$};
  \node[font=\scriptsize, text=black!60, align=center, anchor=north]
        at ([yshift=-1.5mm]rne.south) {rank 16, same on $W_2$};
\end{scope}
\node[ttl] at ([yshift=1mm]B.center |- A.north) {(b) Language-gated ConvNeXt1D block};

\begin{scope}[shift={(14.0,2.55)}, local bounding box=C]
  \node[io] (cm) at (0,0) {log-mel $\in\mathbb{R}^{T\times 80}$};
  \node[dec, sm, above=3mm of cm] (cs) {Conv1D $80\!\rightarrow\!192$};
  \node[dec, sm, above=2.4mm of cs, minimum height=6.4mm] (cx) {ConvNeXt1D $\times 3$, $d{=}192$};
  \node[dec, sm, above=2.4mm of cx] (ch) {Linear $\rightarrow 2{\times}513$};
  \node[dec, sm, minimum width=10mm, above left=2.6mm and -9.5mm of ch] (mag) {$\exp(\cdot)$};
  \node[dec, sm, minimum width=10mm, above right=2.6mm and -9.5mm of ch] (ph) {$\cos,\sin$};
  \node[op, above=8.6mm of ch] (mul) {$\times$};
  \node[dec, sm, above=2.6mm of mul] (ist) {iSTFT (1024 / 240)};
  \node[io, above=2.4mm of ist] (cw) {waveform, 24\,kHz};
  \foreach \a/\b in {cm/cs, cs/cx, cx/ch, mul/ist, ist/cw} \draw[flow] (\a) -- (\b);
  \draw[flow] (ch.north) ++(-5mm,0) -- (mag.south);
  \draw[flow] (ch.north) ++(5mm,0) -- (ph.south);
  \draw[flow] (mag.north) |- (mul.west);
  \draw[flow] (ph.north) |- (mul.east);
  \node[font=\scriptsize, text=black!55, anchor=east] at ([xshift=-1mm]mag.west) {$|X|$};
  \node[font=\scriptsize, text=black!55, anchor=west] at ([xshift=1mm]ph.east) {$\angle X$};
\end{scope}
\node[ttl] at ([yshift=1mm]C.center |- A.north) {(c) iSTFT waveform head};

\end{tikzpicture}}
  \caption{\textbf{PAWP-TTS architecture.} (a) A deterministic frontend (dashed, no parameters) maps mixed Devanagari/Latin text to language-tagged phonemes; every learned inference tensor is counted, including the waveform decoder (44.6\% of parameters). Durations scale token states to 10\,ms frames; predicted pitch and energy condition the acoustic network. (b) Both pointwise layers of every encoder and acoustic block add rank-16 English and Nepali residuals selected per token by the frontend's language mask, enabling code-mixing; for monolingual export they fold into $W$. (c) A Vocos-style head predicts log-magnitude and phase for a 513-bin spectrum and inverts it with an iSTFT.}
  \label{fig:arch}
\end{figure*}
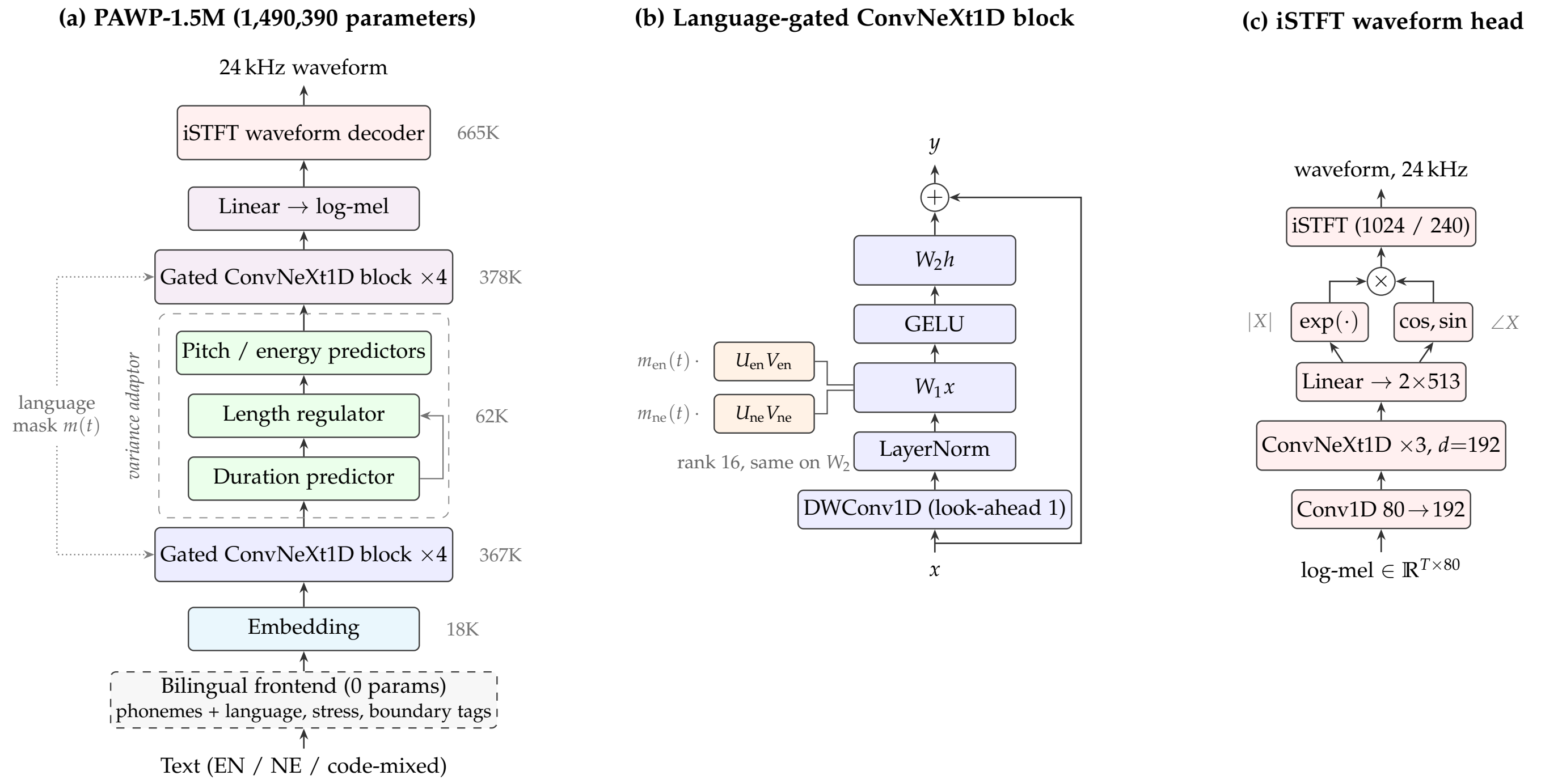

\subsection{Deterministic Bilingual Frontend}
Figure~\ref{fig:arch} gives an overview. Rather than spending neural capacity memorizing orthographic exceptions, PAWP-TTS uses a deterministic frontend mapping mixed Devanagari and Latin text to a unified 128-token IPA inventory with explicit language IDs, lexical stress, and word-boundary (\texttt{<sp>}) markers:
\begin{itemize}
  \item \textbf{Nepali Phonology:} Implements standard Kathmandu Nepali rules~\cite{khatiwada2009nepali} covering the Nepali inherent vowel, alveolar affricates, sibilant merger to \texttt{/s/} (0.0\% Hindi postalveolar \texttt{/sh/} symbol bleed), post-vocalic retroflex flapping, gemination, and a 105,868-word Nepali lexicon ($\approx$99\% rule/paradigm-generated, 100\% symbol-compliant, 0\% \texttt{<unk>}) plus 76 context-dependent heteronyms (152 readings).
  \item \textbf{English G2P:} In-process \texttt{libespeak-ng} via \texttt{ctypes} ($<$0.20\,ms/sentence), seamlessly interleaved with Nepali tokens for Devanagari--Latin code-mixing.
\end{itemize}

\subsection{Language-Gated Superposition and Folding}
Each pointwise projection in the encoder and acoustic blocks is parameterized as:
\begin{equation}
  y_t = W_{\text{shared}} x_t + m_{\text{en}}(t)\,\Delta W_{\text{en}} x_t + m_{\text{ne}}(t)\,\Delta W_{\text{ne}} x_t,
\end{equation}
where $\Delta W_{\ell} = g_\ell U_\ell V_\ell$ are compact low-rank language residuals and $m_\ell(t) \in [0,1]$ are deterministic frontend language masks. During code-mixed synthesis, token- and frame-level masks route smoothly across script boundaries; for monolingual deployment, $\Delta W_\ell$ folds statically into $W_{\text{shared}}$ at export time for zero runtime overhead.

\subsection{Decoder-Centric 1.5M and 281K Tiers}
Two primary parameter tiers are evaluated (Table~\ref{tab:params}):
\begin{itemize}
  \item \textbf{PAWP-1.5M:} 4 encoder ConvNeXt1D blocks ($d=128, k=5$, lookahead 1), variance predictors (duration, $F_0$, voicing, energy), 4 acoustic blocks ($d=128, k=7$), and a 3-layer iSTFT decoder ($d=192$, 513 STFT bins, 24\,kHz, $\text{hop}=240$). Total learned parameters: \textbf{1,490,390} (5.98\,MB FP32).
  \item \textbf{PAWP-300K (Shared vs.\ Dense):} Decoder-centric allocation (52.6\% of parameters in the 24\,kHz iSTFT decoder, $n_{\text{fft}}=512$, 28-dim articulatory conditioning). The shared-basis variant (\texttt{EXP-015b}) pools $K=2$ bases across depth (\textbf{281,078} unique / 328,158 materialized params, 20.96\,MMAC/s); the equal-budget dense baseline (\texttt{EXP-015c}) removes basis sharing (\textbf{281,442} unique = materialized params).
\end{itemize}

\section{Experimental Setup and Results}
\label{sec:exp}

\paragraph{Data and training.}
Models train on about 225\,h: 77\,h of English read speech (a LibriSpeech~\cite{panayotov2015librispeech} subset), 108\,h of Nepali (internally prepared studio recordings and read speech, including OpenSLR Nepali), and 34\,h of code-mixed speech. Utterances are 1--12\,s, and sentences in validation or test splits are removed from training. For a single cross-lingual target voice, a fine-tuned OmniVoice~\cite{omnivoice2026} teacher synthesized English and Nepali in one voice (22.7\,h EN, 15.7\,h NE), kept only when an independent ASR transcribed it with CER $\le$0.10 (EN) or $\le$0.12 (NE). Supplementary synthetic Nepali receives half the loss weight of human recordings. Training is multi-speaker (1,815 speakers); exported models keep one speaker vector. Training runs in BF16 on one RTX~4090 with L1 log-mel, multi-resolution STFT, duration, $F_0$/voicing and energy losses plus multi-period/multi-resolution adversarial and feature-matching losses. On half of the steps the acoustic network is conditioned on its own predicted prosody rather than ground truth, which closed a large train/inference gap. Durations come from Viterbi forced alignment with a training-only CTC phoneme recognizer~\cite{graves2006ctc}.

\subsection{Clean \texttt{v3} Data Alignment (\texttt{EXP-014})}
\begin{figure}[t]
  \centering
  \includegraphics[width=\columnwidth]{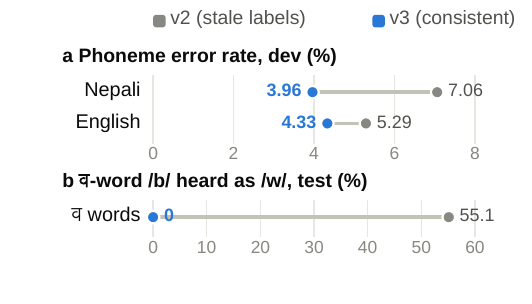}
  \caption{\textbf{Label consistency matters more than any edit.} Rebuilding the feature cache and phoneme recognizer with the current frontend (v2$\to$v3) nearly halves Nepali PER (a) and removes the spurious \textit{va}-word /b/$\to$/w/ errors that earlier looked like a model defect (b).}
  \label{fig:labels}
\end{figure}

Early training (\texttt{v2}) predated the final Nepali lexicon and \textit{va} $\rightarrow$ \texttt{/b/} rule update. Rebuilding the 136,807-utterance feature cache (\texttt{v3}) with the current frontend, retraining the frozen CTC phoneme recognizer (\texttt{ctc\_aligner\_v3.pt}, CTC loss $\approx$1.0 $\rightarrow$ 0.614), and re-extracting Viterbi forced-alignment durations reduced single-frame duration collapses by $\approx$3$\times$ and dropped spurious \textit{va}-word \texttt{/b/ $\rightarrow$ /w/} confusion on 300 Nepali test utterances from 55.1\% (65/118) to \textbf{0.0\%} (0/118). Fine-tuning on consistent labels then lowered validation PER from 7.06\% to 3.96\% (Nepali) and 5.29\% to 4.33\% (English; Figure~\ref{fig:labels}), measured on the in-training validation subset (Section~\ref{sec:wsp} uses a separate fixed 150+150 validation subset). This intervention is confounded: the labels, the recognizer, the durations and 10k additional training steps all changed together, with no continued-\texttt{v2} control. Because the recognizer that scores PER was retrained on the new labels, the 7.06\% and 3.96\% figures are not on a common scale; part of the drop reflects agreement with the new label conventions. The /b/$\to$/w/ result is a direct substitution count and is the more robust evidence.

\subsection{Bilingual Synthesis Quality and Equal-Budget 281K Ablation}
Table~\ref{tab:params} summarizes the parameter footprint and held-out evaluation across \texttt{PAWP-1.5M v3} (\texttt{EXP-015a}, 10k FT steps on \texttt{v3} from \texttt{EXP-012}) and the equal-budget 281K variants (\texttt{EXP-015b} Shared vs.\ \texttt{EXP-015c} Dense, trained from scratch for 36k steps on \texttt{v3}, identical recipe, \texttt{seed=42}). Held-out ASR is measured on $n=30$ EN, $n=30$ NE, and $n=9$ code-mixed utterances using a single recognizer (reported alongside the human studio audio floor as an intelligibility proxy, not a naturalness comparison). Untouched test PER is measured on $n=300$ NE + $n=300$ EN utterances via \texttt{ctc\_aligner\_v3.pt}, and minimal-pair contrast is scored acoustically on an internal 77-pair set (11 pairs in each of seven categories: four-way laryngeal, dental/retroflex, alveolar affricates, gemination, schwa deletion/flapping, sibilant merger/nasality, and English/code-mix contrasts). Both members of a pair are synthesized and decoded by the CTC recognizer. \emph{Distinct} is the fraction of pairs whose two decodings differ, and \emph{Exact} is the fraction where both decode exactly to their frontend phoneme strings.

\begin{figure}[t]
  \centering
  \includegraphics[width=\columnwidth]{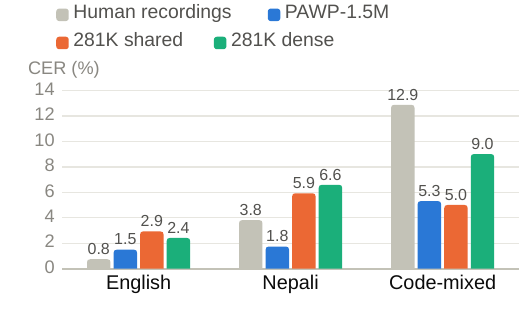}
  \caption{\textbf{Held-out ASR character error rate} (one recognizer; $n=30$ EN, 30 NE, 9 code-mixed sentences). Gray bars are the same recognizer on the human recordings of those sentences; this recognizer also filtered the training corpus, and below-human synthetic CER does not hold under an independent recognizer (Table~\ref{tab:indep}).}
  \label{fig:asr}
\end{figure}

\paragraph{Evaluation validity.}
Both automatic metrics have known circularity. (i)~PER is scored by \texttt{ctc\_aligner\_v3.pt}, the same recognizer that produced the training durations, trained on the same frontend labels. PER therefore measures consistency with the frontend's phoneme conventions, not absolute pronunciation correctness, and may miss distortions that stay inside the recognizer's decision boundaries. (ii)~The ASR service that scores held-out WER/CER is the same recognizer that filtered the synthetic target-voice corpus, which favours speech that this recognizer transcribes well. This explains why synthetic CER can fall below the human-recording floor (Figure~\ref{fig:asr}). Table~\ref{tab:indep} therefore re-scores all models with two recognizers that played no part in training or data filtering. Native-listener tests remain pending.

\paragraph{Independent re-scoring.}
Under independent recognizers (Table~\ref{tab:indep}), every synthetic system is \emph{worse} than the human recordings on English CER and on phone error rate in both languages, so the earlier below-human CER was a recognizer artifact. The independent ASR is weak on Nepali (human CER 22.65\%), and the phone recognizer was not trained on Nepali (human PER 34.12\%). Absolute Nepali numbers are therefore only coarse, and the comparisons below are paired on identical sentences. PAWP-1.5M remains close to the human reference on English (PER 15.14\% vs.\ 13.44\%; CER 2.35\% vs.\ 0.68\%). On a new 160-sentence code-mixed set (4.7 language switches per sentence, no textual overlap with training data, zero \texttt{<unk>} tokens), it reaches 18.13\% WER / 8.80\% CER (95\% CI 15.9--20.6\% WER), with more errors in English spans (21.9\%) than in Nepali spans (12.0\%). The phone recognizer also outputs no retroflex nasal on any of the 300 human Nepali utterances, so it can neither confirm nor refute the nasal finding of Section~\ref{sec:wsp}.

\begin{table*}[t]
\centering
\caption{End-to-end parameter footprint (all inference tensors counted) and evaluation on clean \texttt{v3} alignment. Held-out ASR uses a single recognizer ($n=30/30/9$, \texttt{seed=42}); untouched test PER uses \texttt{ctc\_aligner\_v3.pt} ($n=300$ NE + $300$ EN), reporting 3-seed mean $\pm$ std (\texttt{seeds 42, 101, 202}) for \texttt{EXP-015b/c}.}
\label{tab:params}
\footnotesize
\setlength{\tabcolsep}{4pt}
\resizebox{\textwidth}{!}{%
\begin{tabular}{@{}lrrrrrrr@{}}
\toprule
Model (\texttt{v3}) & Unique / Total & FP32 & EN ASR & NE ASR & Mix ASR & Test NE / EN & CTC 77-Pair \\
 & Params & Size & WER/CER & WER/CER & WER/CER & PER ($n=600$) & Distinct / Exact \\
\midrule
Human Floor (ASR) & --- & --- & 2.63 / 0.75 & 11.60 / 3.82 & 25.91 / 12.88 & --- & --- \\
\textbf{PAWP-1.5M v3} (\texttt{EXP-015a}) & 1,490,390 / 1,490,390 & 5.98\,MB & \textbf{4.02 / 1.50} & \textbf{12.35 / 1.75} & 12.32 / 5.32 & \textbf{5.04\% / 4.02\%} & \textbf{83.1\% / 58.4\%} \\
\textbf{PAWP-300K Shared} (\texttt{EXP-015b}) & 281,078 / 328,158 & 1.12\,MB & 6.88 / 2.93 & \textbf{19.71 / 5.93} & \textbf{12.02 / 5.01} & 6.35$\pm$0.11\% / 6.57$\pm$0.47\% & 77.9\% / 39.0\% \\
\textbf{Dense-300K} (\texttt{EXP-015c}) & 281,442 / 281,442 & 1.13\,MB & \textbf{6.15 / 2.42} & 20.21 / 6.59 & 18.09 / 9.01 & \textbf{5.53$\pm$0.08\% / 5.34$\pm$0.28\%} & 79.2\% / 37.7\% \\
\bottomrule
\end{tabular}}
\end{table*}

\begin{figure}[t]
  \centering
  \includegraphics[width=\columnwidth]{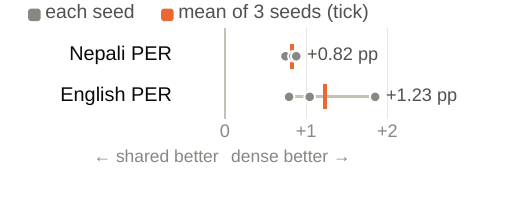}
  \caption{\textbf{Shared-basis vs.\ dense at equal unique parameters (281K).} Difference in untouched-test PER (shared $-$ dense; $n=300$ NE + 300 EN) for three training seeds (gray) and their mean (orange). Dense is better on every seed in both languages.}
  \label{fig:shared}
\end{figure}

\paragraph{Multi-Seed Shared vs.\ Dense Trade-Off.}
At \texttt{seed=42}, \texttt{PAWP-300K Shared} achieved lower \texttt{val\_mel} (0.8768 vs.\ 0.8778), stronger \texttt{val\_energy} (1.50 vs.\ 1.60), and lower Code-Mix ASR WER (12.02\% vs.\ 18.09\%, $n=9$; not reproduced at $n=160$, see below). However, on the larger untouched $300+300$ \texttt{test.jsonl} PER benchmark, equal-unique-parameter \texttt{Dense-300K} outperformed \texttt{Shared} on \textbf{3 out of 3 independent seeds} (42, 101, 202) with zero overlap: Nepali PER averaged \textbf{5.53$\pm$0.08\%} vs.\ 6.35$\pm$0.11\% ($-0.82$\,pp; \texttt{seed=42} paired bootstrap 95\% CI $[+0.56, +1.41]$, $p<0.001$) and English PER averaged \textbf{5.34$\pm$0.28\%} vs.\ 6.57$\pm$0.47\% ($-1.23$\,pp; \texttt{seed=42} 95\% CI $[+1.44, +2.35]$, $p<0.001$), approaching \texttt{PAWP-1.5M v3} at 5.3$\times$ fewer parameters (Figure~\ref{fig:shared}). The confidence intervals capture utterance sampling within seed 42 only. Across seeds, the evidence is the sign of the difference (dense better in 6/6 seed$\times$language comparisons) and the non-overlapping seed ranges. Other metrics disagree: on the small held-out ASR set (seed 42), \texttt{Shared} has lower Nepali CER (5.93 vs.\ 6.59) and code-mix CER (5.01 vs.\ 9.01, $n=9$). Independent re-scoring resolves most of this disagreement. Under the independent phone recognizer, dense again wins all 6 seed$\times$language comparisons (EN 14.83$\pm$0.32\% vs.\ 16.28$\pm$1.14\%, NE 38.69$\pm$0.62\% vs.\ 40.51$\pm$0.42\%; paired $t$-test over the six pairs $p=0.010$, Wilcoxon $p=0.031$) and all 3 seeds on the code-mixed set (36.49\% vs.\ 39.47\% PER). Independent English ASR favours dense on all 3 seeds (WER 5.22\% vs.\ 8.13\%), while Nepali ASR is tied. The seed-42 code-mix advantage of \texttt{Shared} disappears on the 160-sentence set (18.78\% WER for both). The conclusion: at equal unique parameters, $K=2$ basis sharing gives no advantage, and dense is better on phone accuracy under both the training-time and an independent recognizer. Note that \texttt{Shared} materializes 17\% more compute parameters (328K vs.\ 281K), so a compute-matched comparison would favour dense further. Architectures: \texttt{Dense-300K} uses $d_{\text{model}}=58$, 3 encoder and 3 acoustic ConvNeXt blocks (FFN 116, kernels 5/7), rank-4 language residuals, and a 3-block decoder of width 88 (FFN 176) with a 257-bin iSTFT head ($n_{\text{fft}}=512$). \texttt{Shared} uses $d_{\text{model}}=64$ (FFN 128), $K=2$ shared bases with rank-8 per-layer residuals, 28-dim articulatory conditioning, and decoder width 96 (FFN 192). Both use duration/pitch/energy predictors of width 48/48/32.

\begin{table}[t]
\centering
\caption{\textbf{Independent re-scoring} (\texttt{EXP-018/019}). ASR: a second mixed-script recognizer not used for data filtering (held-out $n=30$ EN / 30 NE). PER: public \texttt{wav2vec2-xlsr-53-espeak-cv-ft} phone recognizer, both sides mapped to a shared 39-phone set (test $n=300$ EN / 300 NE). Mix: new 160-sentence code-mixed set (1,633 words), micro-averaged. 281K rows: mean$\pm$std over seeds 42/101/202.}
\label{tab:indep}
\footnotesize
\setlength{\tabcolsep}{2.5pt}
\resizebox{\columnwidth}{!}{%
\begin{tabular}{@{}lrrrrr@{}}
\toprule
 & EN & NE & EN & NE & Mix-160 \\
Model & CER & CER & PER & PER & WER / CER \\
\midrule
Human & 0.68 & 22.65 & 13.44 & 34.12 & --- \\
PAWP-1.5M & 2.35 & 19.46 & 15.14 & 38.10 & 18.13 / 8.80 \\
Shared-300K & 4.10{\scriptsize$\pm$0.52} & 21.07{\scriptsize$\pm$1.93} & 16.28{\scriptsize$\pm$1.14} & 40.51{\scriptsize$\pm$0.42} & 18.78 / 9.35 \\
Dense-300K & 2.44{\scriptsize$\pm$1.20} & 20.18{\scriptsize$\pm$0.78} & \textbf{14.83}{\scriptsize$\pm$0.32} & \textbf{38.69}{\scriptsize$\pm$0.62} & 18.78 / 9.00 \\
\bottomrule
\end{tabular}}
\end{table}

\paragraph{Latency and Zero-Shot Compression.}
On a 4-thread x86\_64 CPU (30 timed runs after 3 warm-up runs), the folded monolingual \texttt{PAWP-1.5M} ONNX graph computes its first 60\,ms audio chunk in 1.11\,ms p50 (1.52\,ms p99), i.e.\ 1.9\% of the chunk duration, and the frontend takes 0.16\,ms p50 per sentence. Over whole utterances, the measured real-time factor is 0.0058. Latency depends only on graph shape, so it was measured on the structural (untrained) graph; these are compute times measured after the utterance is encoded. In a later rerun of the trained graphs under the same protocol, the code-mixed graph (residuals unfolded, both languages active) took 1.24\,ms p50 per 60\,ms chunk vs.\ 0.96\,ms folded (+29\%; RTF 0.0055 vs.\ 0.0043). The 281K models took 0.70\,ms folded and 0.82--0.84\,ms unfolded (+17--19\%). Not yet measured: end-to-end time-to-first-audio with stateful streaming, and ARM/browser targets. Post-hoc one-shot SVD of the trained 1.5M model without recovery training (\texttt{EXP-010}; held-out ASR, $n=30/30/9$) raised English WER from 5.0\% to 14.3\% at 1.0M (uniform SVD) and to 66\% at 750K, and every method collapsed (WER $>100\%$) at 500K. The trained-from-scratch 281K models avoid this collapse, which suggests that compact end-to-end training is needed below $\sim$1M parameters.

\section{Diagnostic Weight-Space Programming: Mechanisms and Confounders}
\label{sec:wsp}

\begin{figure*}[t]
  \centering
  \includegraphics[width=0.8\textwidth]{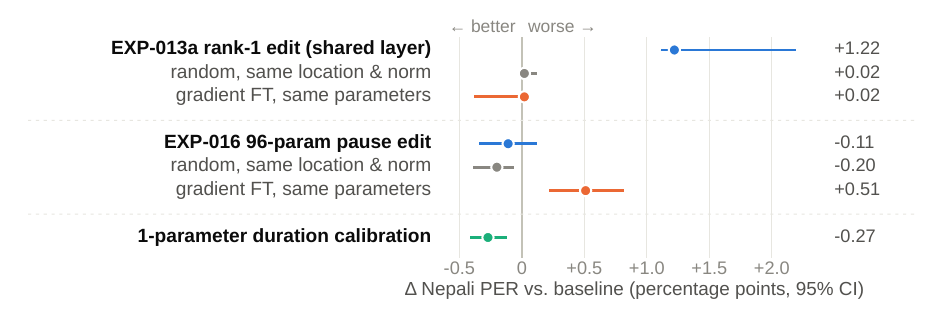}
  \caption{\textbf{Weight edits against matched controls.} Mean paired per-utterance change in Nepali PER on untouched test data ($n=300$, paired bootstrap 95\% CI), each relative to the unedited model in the same scoring pass. Edits (blue) are compared with a random edit of the same location and norm (gray) and gradient fine-tuning of the same parameters (orange). The unconstrained rank-1 edit regresses; the pause edit is not better than its random control; a single global duration-bias parameter (green), selected on validation, gives the only significant improvement.}
  \label{fig:forest}
\end{figure*}

Zero-training, activation-guided weight edits were evaluated systematically on \texttt{PAWP-1.5M} under strict controls: hyperparameter selection on \texttt{val.jsonl} only, one-shot verification on untouched $300+300$ \texttt{test.jsonl}, paired-bootstrap 95\% CIs, matched-norm random edits, and same-parameter gradient fine-tuning controls. Three experiments revealed both genuine mechanistic properties and critical methodological pitfalls; Figure~\ref{fig:forest} summarizes every edit against its controls. Because several edits and controls were tested and $p$-values are not corrected for multiple comparisons, all results in this section are exploratory. Edits are ROME-style~\cite{meng2022rome} rank-1 updates. For a linear map $W$, let $\bar h$ be the mean input activation over validation frames of the target class and $C$ the input covariance over non-target frames. The key direction is $\hat k = C_\lambda^{-1}\bar h / \lVert C_\lambda^{-1}\bar h\rVert$ with $C_\lambda = C + \lambda I$, so the edit responds weakly to non-target inputs, and the update is $\Delta W = \alpha\,\lVert W\rVert_F\,\hat u\,\hat k^\top$, where $\hat u$ is the unit output direction to be strengthened and the scale $\alpha$ is swept on validation only. Random controls use Gaussian unit vectors $\hat u_r,\hat k_r$ at the same location with the same $\alpha\lVert W\rVert_F$. Gradient controls fine-tune exactly the same parameters with Adam, with learning rate and step count also selected on validation.

\subsection{Unconstrained Shared Edits vs.\ Language-Residual Isolation}
In \texttt{EXP-013a}, an unconstrained Rank-1 edit in shared \texttt{acoustic\_blocks.3.pw2.weight} reduced target confusion but degraded overall PER (NE $+1.22\%$, EN $+1.75\%$). Furthermore, grapheme auditing revealed that 90.3\% (65/72) of the targeted \texttt{/b/ $\rightarrow$ /w/} errors were a \texttt{v2} cache-label artifact on \textit{va} words. Rebuilding \texttt{v3} eliminated the artifact entirely, underscoring that mechanistic editing must never be applied before label consistency is verified.

\subsection{Word-Boundary Pause Selectivity and Global Duration Bias (\texttt{EXP-016})}
On clean \texttt{v3} validation data, word-boundary \texttt{<sp>} tokens were under-predicted ($6.16$ vs.\ $10.38$ frames) and 90.8\% linearly separable from regular phonemes at the 96-dim input of \texttt{duration\_predictor.proj.weight} (96 params = 0.0064\% of model). A ROME-whitened Rank-1 update achieved \textbf{27:1 selectivity} on untouched test data (\texttt{<sp>} $+1.001$ frames vs.\ phonemes $+0.037$ frames), reduced deletions ($-19$ NE, $-21$ EN), and beat dev-tuned 96-param gradient FT (NE PER 4.93\% vs.\ 5.55\%; Figure~\ref{fig:forest}).

\begin{figure*}[t]
  \centering
  \begin{subfigure}[t]{0.47\textwidth}
    \centering
    \includegraphics[width=\linewidth]{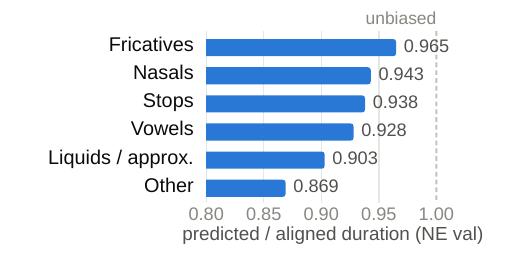}
    \caption{Predicted / aligned duration by phoneme class.}
  \end{subfigure}\hfill
  \begin{subfigure}[t]{0.47\textwidth}
    \centering
    \includegraphics[width=\linewidth]{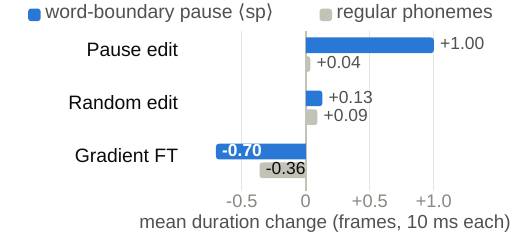}
    \caption{Duration change from the 96-parameter pause edit and controls.}
  \end{subfigure}
  \caption{\textbf{Duration diagnostics.} (a) Every phoneme class is predicted 3.5--13\% too short, a global bias consistent with log-domain duration training. (b) The pause edit lengthens word-boundary pauses 27$\times$ more than regular phonemes, but a random edit of equal norm also lengthens everything slightly, which explains its PER gain.}
  \label{fig:dur}
\end{figure*}

However, the edit did not beat a matched random-96 control on overall NE PER (4.93\%, $p=0.177$ vs.\ 4.84\%, $p=0.004$). A per-class duration audit (all 169 Nepali utterances of the validation split, 13,450 phonemes; the PER sweep below uses a fixed 150+150 subset) found the cause: baseline durations are \textbf{globally biased short by $-7.3\%$} across every phoneme class (pred $6.105$f vs.\ GT $6.585$f; ratio $0.927$ overall, ranging from $0.869$ to $0.965$), consistent with Jensen's inequality under log-duration training. Selecting a 1-parameter global duration bias shift (\texttt{duration\_predictor.proj.bias}) strictly on \texttt{val.jsonl} ($n=150$ NE + $150$ EN; selected $\Delta b = +0.10$, improving average validation PER from 4.625\% to 4.448\%) and evaluating \textbf{once} on untouched \texttt{test.jsonl} ($n=300$ NE + $300$ EN) reduced corpus-level Nepali PER from 4.744\% ($D=401$) to \textbf{4.513\%} ($D=373$, $-0.23$\,pp); the mean paired per-utterance change is $-0.27$\,pp (95\% CI $[-0.41, -0.12]$, $p<0.001$), with no significant English change (4.044\% $\rightarrow$ 4.016\%, $p=0.317$). This audit used its own synthesis and scoring pass, whose unedited baseline (4.744\% NE, 4.044\% EN) differs from the Table~\ref{tab:params} pass (5.04\% / 4.02\%) by up to 0.29\,pp. All deltas in this section and in Figure~\ref{fig:forest} are therefore paired within a single pass, and the pass-to-pass variation is itself of the same order as the effects studied. Global speaking-rate calibration thus explains the random-96 gain. Because the predictor outputs linear frames but is trained with a log-domain loss, a $+0.10$ pre-softplus shift adds only $\approx$0.1 frame per phoneme and \emph{partially} corrects the $-7.3\%$ bias; the proper fix is a bias-corrected duration loss at training time.

\subsection{Target Validity Against Native Human Recordings (\texttt{EXP-017})}
After \texttt{v3} re-alignment, the top remaining Nepali substitution on validation was retroflex nasal to dental nasal (IPA retroflex nasal $\rightarrow$ dental /n/). Applying a ROME null-space Rank-1 edit strictly inside the Nepali-only low-rank residual \texttt{encoder\_blocks.0.pw2.\{u\_ne, v\_ne\}} (6,144 params = 0.41\%) left English output unchanged (by construction, since the residual is gated off for English, so this is an architectural guarantee rather than a measured property of the edit) and changed overall Nepali PER by only 0.01\,pp (5.036\% $\rightarrow$ 5.026\%, $p=0.495$). The edit also failed on its target. With the validation-selected scale ($\alpha=0.01$, $\lVert\Delta W\rVert_F=0.259$), the target confusion rate on test was unchanged (41.07\%, $n=56$ occurrences), and the target logit margin moved by only $-0.009$ (95\% CI $[-0.028, +0.006]$, $p=0.148$). Same-parameter gradient fine-tuning moved the margin significantly more ($-0.041$, $p<0.001$), although it too left the confusion rate unchanged.

Critically, running \texttt{ctc\_aligner\_v3.pt} directly on \textbf{ground-truth human studio Nepali audio} in \texttt{test.jsonl} ($n=62$ occurrences across 52 utterances) decoded \textbf{90.3\% (56/62)} of orthographic retroflex-nasal occurrences as dental /n/ (mean margin $+3.246$). This measurement has a caveat: the recognizer was itself trained on frontend labels and may not discriminate the retroflex--dental nasal contrast reliably, and the reference inventory~\cite{khatiwada2009nepali} lists a retroflex nasal. Confirmation by native listeners is therefore pending. On the current evidence, \texttt{EXP-017}'s target is better explained as a frontend phonology choice (proposed as an unimplemented future \texttt{v4} frontend rule, as ~1,800 lexicon entries still emit the retroflex symbol), not a neural model defect.

\section{Limitations and Conclusion}
\label{sec:conclusion}

\paragraph{Limitations.}
(1) Evaluation is automatic only: no listening test (MOS, intelligibility or A/B) has been run, which is the highest-priority future work. The training-time metrics are partly circular. The independent recognizers used to address this are weak on Nepali (human CER 22.65\%, human PER 34.12\%), so Nepali comparisons rest on paired differences, not absolute values. The monolingual held-out ASR sets are small ($n=30$ per language).
(2) The 105,868-word Nepali lexicon is $\approx$99\% rule/paradigm-generated and has not yet been audited against an independent native-annotated gold set; the proposed orthographic retroflex-nasal $\rightarrow$ dental /n/ mapping is documented as a candidate \texttt{v4} frontend update and is not yet implemented in the released \texttt{v3} weights.
(3) Models synthesize a single target speaker identity at inference time without zero-shot voice cloning.
(4) The target-voice corpus is teacher-generated; perceptual effects of distilling from a synthetic voice are not yet measured.
(5) Latency is reported as per-chunk compute on one x86 CPU; end-to-end time-to-first-audio with stateful streaming and ARM/browser targets are not yet measured.
(6) There is no grapheme-input or neural-G2P baseline at equal budget, so the frontend's contribution is not isolated from the acoustic architecture; the language-gated residuals are not ablated (no rank sweep or no-residual control); and compact systems such as LightSpeech~\cite{luo2021lightspeech} and Nix-TTS~\cite{chevi2022nixtts} have not been re-counted and compared under the same accounting rule.
(7) The 1.5M model was fine-tuned from an earlier checkpoint, whereas the 281K models were trained from scratch, so cross-tier comparisons are not training-matched. Seed-level confidence intervals are reported for seed 42 only, and the weight-editing results are not corrected for multiple comparisons.
(8) Part of the pipeline is not publicly reproducible: internally prepared Nepali/code-mixed recordings, the teacher TTS model and the primary ASR service. The code, frontend, lexicon, configurations, ONNX graphs, weights, training-time PER recognizer and the 160-sentence code-mixed test set are released.

\paragraph{Conclusion.}
This work demonstrated that combining a linguistically grounded English--Nepali frontend with a compact, decoder-centric ConvNeXt1D+iSTFT backbone yields intelligible 24\,kHz bilingual and code-mixed speech at 1.49M parameters (5.98\,MB) and 281K parameters (1.12\,MB). Equally important, the controlled WSP audits establish practical methodological guardrails for weight editing in speech synthesis: verify label consistency, validate target errors directly on human audio, isolate edits to language-specific low-rank residuals, and benchmark against 1-parameter global calibration and matched-budget gradient controls.

\paragraph{Acknowledgments.}
This is independent research by the author. The author thanks WiseYak for providing GPU compute and for access to (i)~internally prepared Nepali and code-mixed speech recordings used in training, (ii)~an internally fine-tuned text-to-speech model, used only through inference to synthesize the target-voice corpus, and (iii)~an internal speech-recognition service used for evaluation. All other training data are public corpora (Section~\ref{sec:exp}). The weight-space programming experiments (Section~\ref{sec:wsp}) were designed and run with the assistance of an AI research agent (Gemini~4 Neon), which proposed edits and ran experiment scripts. Independent AI agents audited the methodology and results; this audit led to the corrections reported in Section~\ref{sec:wsp} (the label artifact, the duration-bias control and the nasal-target re-check). Every reported number was checked against logged result files, which are released with the code. The author designed the study, reviewed all results and is responsible for the content.

\bibliographystyle{plain}
\bibliography{references}

\end{document}